\documentclass{article}

\usepackage[nonatbib]{neurips_2020}

\usepackage[utf8]{inputenc} % allow utf-8 input
\usepackage[T1]{fontenc}    % use 8-bit T1 fonts
\usepackage{hyperref}       % hyperlinks
\usepackage{url}            % simple URL typesetting
\usepackage{booktabs}       % professional-quality tables
\usepackage{amsfonts}       % blackboard math symbols
\usepackage{nicefrac}       % compact symbols for 1/2, etc.
\usepackage{microtype}      % microtypography
\usepackage[dvipsnames]{xcolor}
\usepackage{graphicx}
\usepackage[ruled,vlined]{algorithm2e}
\usepackage{textcomp}

\title{Placement in Integrated Circuits using Cyclic Reinforcement Learning and Simulated Annealing}

\author{
}

\begin{document}

\maketitle

\begin{abstract}
% 1. Succinct statement of problem and why we should care\\
% 2. Clear statement of what you are doing and how your technique or approach solves this problem\\ 
% 3. Key details of your technical solution\\
% 4. Results of experiments evaluating your solution, or the most significant takeaways of your work
% Heuristic Algorithms follow a hill climbing approach to come up with an optimal solution. This approach is largely dependent on their point of initialisation. One such Heuristic algorithm called Simulated Annealing is widely used for placing circuit components on an Integrated Circuit. Improving such a widely used algorithm is of great value to the IC placement process. In this paper, we devise a Cyclic Reinforcement Learning framework to improve the solution generated by Simulated Annealing on the cost of increasing its run time. Our framework helps the agent to generate a better initialisation point by cyclically reducing the difference between the cost of its solution and the one generated by Simulated Annealing. Results show that the agent is able to produce more optimal solutions than randomly initialised Simulated Annealing. To the best of our knowledge, we are the first to apply such a Reinforcement Learning framework to the domain of IC placement.
% \textbf{need to finalize}
Physical design and production of integrated circuits (IC)  is becoming increasingly more challenging as the sophistication in IC technology is steadily increasing. Placement has been one of the most critical steps in IC physical design. Through decades of research, partition-based, analytical-based and annealing-based placers have been enriching the placement solution toolbox. However, open challenges including long run time and lack of the ability to generalize continue to restrict wider applications of existing placement tools. We devise a learning-based placement tool based on cyclic application of reinforcement learning (RL) and simulated annealing (SA) by leveraging the advancement of RL. Results show that the RL module is able to provide a better initialization for SA and thus leads to a better final placement design. Compared to other recent learning-based placers, our method is majorly different with its combination of RL and SA by leveraging the RL model’s ability to quickly get a good rough solution after training and the heuristics' ability to realize greedy improvements in the solution.

\end{abstract}

\section{Introduction}
With Integrated Circuits (IC) becoming increasingly sophisticated following the Moore's Law, IC system design is becoming more challenging. Electronic Design Automation (EDA) tools play a dominant role in addressing IC design challenges, especially in advanced node technology IC. Physical IC design is one of the most critical and involved steps in EDA tools. It determines the quality of the final results. Within physical design, placement is one of the most time-consuming and challenging stages. The placement step consists of placing a set of instances from a netlist to a chip canvas so that objectives related to the performance such as power, performance and area (PPA) can be minimized, while satisfying constraints such as routing congestion and density [1]. 
% Practically, when evaluating the performance of placers, approximated costs such as Half Parameter WireLength (HPWL) [2] or chip areas are used. 

% (Status quo) 
Due to its critical role in IC design, placement has been extensively researched for decades. Most existing placement solvers can be classified into the following three categories: partition-based, analytical-based and annealing-based [2].
% These placers are usually based on either heuristics [5] or quadratic optimization [3,4].
Early partition-based placers [14,15] were featured with divide-and-conquer ideas in their algorithms, later ones were based on multi-level partitioning methods [16]. The structure of partition-based placers enables them to solve larger size problems. However, this comes at the cost of solution quality as they are prone to losing global optimality. Analytical-based placers, as the second category, have also been developed over decades. These placers mostly depend on optimizations such as quadratic methods and non-linear optimizers. Some of the more recent and advanced placers are, in fact based on analytical methods such as ePlace [17] and RePlAce [18]. The analytical methods, however, are only restricted to problems that have objectives and constraints or at least their approximations that are both analytical and differentiable. The third category of placers which are based on annealing methods, have also been researched extensively over the last few decades. In the annealing-based methods, Simulated Annealing (SA) or its variations are applied to optimize the placement solution by perturbing potential placement solutions with actions such as swapping and rotation [9,22]. Annealing-based placers are relatively flexible, do not need analytical objectives and constraints, and can find the global optimum. On the other hand, they suffer from long run times and do not accumulate experience over iterations. Modern placers such as Force-Directed Placement [6] were able to achieve incrementally better performance with better problem formulation, better heuristics and optimization techniques such as improved smooth approximation of cost function.Owing to the various problem settings and objectives, there is no universal agreement upon which placers are considered the best and thus the comparison between placers usually depends on implementation and problem formulation. Due to this reason, in our current initial stage of developing learning-based placers, we have chosen benchmarks and baseline algorithms in the annealing-based placers category. The main reason is that annealing-based methods are the most flexible in dealing with different kinds of benchmarks with various constraints, which facilitate our next step of testing the generalization ability of our placer. Also, annealing-based baseline algorithms do not lose global optimality compared with partition-based methods, which make them fair baselines

Even with all the existing placement tools, there are still open challenges in placement. Firstly, one of the key challenges is the long run time of placers. It usually takes experienced electrical engineers weeks to finish a placement design for a IC design even with advanced placers. Secondly, the scalability of existing placers is limited. For instance, recent placers [7,8,9] that use mixed size placement benchmarks such as those from Microelectronics Centers of North Carolina (MCNC) [23,25] can only solve designs with very limited devices ($<600$). The scalability of those placers is, to a great extent, restricted by a lack of better representation of state space for larger size problems. As a consequence, current placers fail to optimize over large state space and thus suffer from the "curse of dimensionality". Other open challenges also include placement with more complex DRCs such as additional geometrics and performance constraints [2].

Recently, learning-based algorithms especially reinforcement learning (RL) have been applied to solve placement aimed at addressing above mentioned challenges [1]. The most important feature that differentiates learning-based placer to previous placers is that the learning-based placer can learn from past experience and improve its performance over training. Besides the learning ability, there are also two other promising properties of learning-based placers. Firstly, the trial-and-error mechanism of RL makes it easier to deal with complex design rule constraints (DRC). Secondly, machine learning models in general tend to have a better state representation ability with flexible representation forms such as image (matrix), vector and graph, which is usually not available for heuristics-based or optimization-based methods. Prior work [1] has symbolized a successful attempt in solving placement with RL, showing the model's ability to generalize across different designs. However, based on related works in RL for solving Combinatorial Optimization (CO) problems similar to placement [10, 11,12], end-to-end RL methods might not usually be the best approach. Combining RL with some other existing algorithms including search or heuristics tends to improve the RL model performance by leveraging the RL model's ability to get a rough good solution fast and the heuristics ability to realize greedy improvement of solution by local adjustment.

% (Gap) 

\begin{figure}
\centering
\includegraphics[width=1\textwidth]{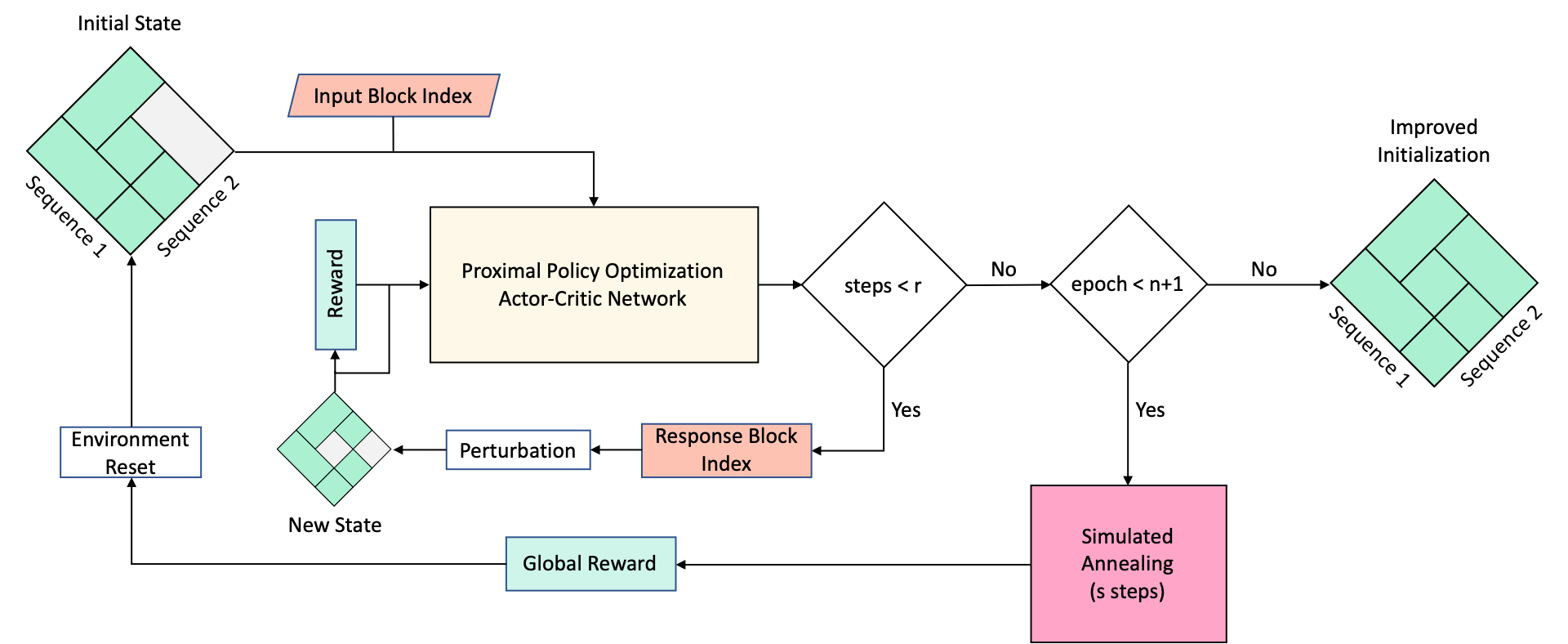}
\caption{Algorithm for cyclic reinforcement learning and simulated annealing for placement}
\label{flow}
\end{figure}

% (Fill the gap) 

% (Contribution) 
Our work is the first step towards solving the placement problem by combining RL and heuristics, inspired by the recent success of solving CO problems with the RL and heuristic algorithms [10]. By implementing and testing our method in different benchmarks, we show that RL is able to provide better initialization for SA and thus improve the final placement solution.

\section{Method}
\label{headings}

\subsection{Problem Description}
We are working towards increasing the efficiency of placement solvers. The goal is to leverage RL for generating a better initialization for SA in solving placement through learning. In this section, we present a synopsis of the general terms used in an IC placement problem, followed by an elaborate account on our RL framework. We define key terminologies and variables in this work as follows:  

\begin{itemize}
    
    \item Sequence pair $P_{seq}$ [19]: a pair of sequences representing the relative spatial relationships between the circuit components (referred to as blocks) by permutations of the blocks' indices. It enforces horizontal and vertical constraints for every pair of blocks by virtue of their relative positions in the Sequence Pairs. The Sequence Pair after $n$ steps of operation is denoted as $P_{seq}(n)$.
    \item Pre-placed block $b_p$: a block whose dimensions and coordinates on the placement plane are specified and fixed.
    \item Free block $b_f$: a block whose dimensions are known but coordinates on the placement plane are not specified. 
    \item Packing $P_{pac}$: a non-overlapping placement of the blocks following a $P_{seq}$, starting at the origin of the placement plane such that all blocks are placed at specified x and y coordinates. For a specific problem, there is a one on one mapping from Sequence Pair to Packing: $P_{seq} \rightarrow P_{pac}$
    \item Chip area $A$: the area of the minimum bounding rectangle of all blocks whose lower left corner is at the origin of the plane describing the placement of the sequence pair.
    \item Wirelength $W$: for a single net, the summation of the Manhattan distances along the $x$ and $y$ directions between blocks that are connected to each other in the net; for a problem, the wirelength is just summation of wirelength for all nets.
    \item Netlist $N$: a set of information including all nets to be placed on an IC design. For individual net, the information includes the indices, dimensions, coordinates and connections of all the blocks in that single net.
    \item Cost $C$: the cost $C$ in this work is the weighted linear combination of the chip area and wirelength.
\end{itemize}

\subsection{Our Approach}
The RL agent learns to generate a Sequence Pair $P_{seq}$ that  serves as a good initialization for SA to further explore the solution space and come up with an optimal solution. We are trying to accomplish this by formulating a cyclic framework between the RL and SA, as illustrated in Fig.~\ref{flow}. The agent interacts with the environment to generate new $P_{seq}$ iteratively. After this, SA will further optimize the $P_{seq}$. This cyclic process is based on the state representation, action representation and reward function of our customized environment specified as follows:
\begin{itemize}
   \item \textbf{State s}: state representation of the environment is a concatenation of the present $P_{seq}$ and a one hot encoding of the index of a randomly chosen block, referred as the \emph{input} block $b_i$. The motivation to select a Sequence Pair based state representation is to enable our framework to generalize to placement problems with non-uniform blocks. It is also much convenient to represent block movements through a Sequence Pair representation.
%   and this representation is further used in Simulated Annealing. 
   \item \textbf{Action a}: action space consists of choosing a \emph{candidate} block $b_c$ from all blocks in response to the current $P_{seq}$ and the \emph{input} block $b_i$. The agent eventually learns to select a \emph{candidate} block such that the $P_{seq}$ generated after a series of swaps between the \emph{candidate} block $b_c$ and the \emph{input} block $b_i$ results in a better initialization for SA.
   \item{\textbf{Reward r}}: the agent takes an action to modify the present $P_{seq}$ and the reward function assigns a reward to the action taken. Here we are referring to the local reward $r_l$ which is the difference between the costs based on $P_{seq}$ before and after the agent's action. The global reward $r_g$ is defined later. We use summation of local and global reward for our RL model.  
\end{itemize}
Our agent is an Actor network that predicts the index of \emph{candidate} block $b_c$ and the value of the action taken by the agent is calculated by the Critic network. The value function, V is further used to compute the advantage of the action taken using Generalised Advantage Estimation[20]. Together, the Actor and Critic networks constitute the policy network, which learns a sequence of actions to provide a good initialization for SA. %maximize the cumulative award.
We use PPO [21] to train the policy network. The RL agent runs \emph r number of steps and $r$ here is a hyperparameter. Now, the Sequence pair generated after \emph r steps is taken as a starting point for SA. The SA runs for another \emph s steps, which is also a hyperparameter.

After $\emph{r} +\emph{s}$ steps in every epoch, a global reward is obtained. We define the global reward $r_g$ as the difference between the costs of solution obtained after running the SA for \emph s steps and the starting point $P_{seq}$ generated after running the RL for \emph {r} steps, which is represented by the following equation:
\begin{equation}
\label{global_reward}
    r_g = C(P_{seq}(r+s)) - C(P_{seq}(r))\hfill\break
\end{equation}

The global reward conveys the efficacy of the  Sequence Pair initialization to the RL as a feedback and serves as the approximated value of the \emph{r}th (final) step V\textsubscript{\emph r}, as stated in  Eqn.~\ref{v_terminal} and Eqn.~\ref{r_g}, in an otherwise infinite action space [10]. Hence, the agent is encouraged to increase the global reward by learning to generate a better initialization $P_{seq}(r)$ for SA after \emph{r} steps of RL operations. The state is reset after every epoch. Since, the weights of our Actor-Critic Network are updated cyclically, we describe the framework as cyclic .

\begin{equation}
\label{v_terminal}
    V_{t}  =\sum_{i=t}^{r-1}\gamma^{i}r_{li}+\sum_{i=r}^{\infty}\gamma^{i}r_{li}  
    =\sum_{i=t}^{r-1}\gamma^{i}r_{li}+V_{r}\hfill \break
\end{equation}

\begin{equation}
\label{r_g}
    r_g =V_{t}-\sum_{i=t}^{r-1}\gamma^{i}r_{li}         
\end{equation}

% \[Reward\textsubscript{ Global} = Cost(Sequence Pair\textsubscript{(r+s)}) - Cost(Sequence Pair\textsubscript{(r)})\]

% \textcolor{red}{what are we leaving this space for? Can we move Preliminary results up?}

\section{Experiments}
\subsection{\textbf{Experimental Setup}}

For the experiments, the PPO algorithm and environment were implemented in Python3.7. All experiments were run on a 1.4GHz CPU together with 8.00GB of memory. The SA part of the algorithm was implemented using the simanneal [24] module.  The experiments were conducted using two different kinds of benchmark problems, the details of which are mentioned below.

\textbf{Benchmarks}. We utilize the lattice problem and the ami49 [23] benchmarks for our experiments. Both the netlists and blocks setting are different. The lattice problems, which are generated with in house developed code, consisting of \emph{n\textsuperscript{2}} blocks of equal width and height where the \emph{i}\textsuperscript{th} block is connected to the (\emph{i}+1)\textsuperscript{th} and the (\emph{i+n})\textsuperscript{th} blocks to form a net. Its cost function is focused on reducing the wirelength of the final placement. Thus, the weight of wirelength is $1$ and the weight of area is $0$ in its cost function. The ami49 problem, on the other hand, has 49 blocks with dimensions varying from 3200 \textmu m to 170 \textmu m (Fig. ~\ref{ami_place}). Its cost function is based on reducing the area of the final placement.Hence, the weight of area is $1$ and the weight of wirelength is $0$ in its cost function.

%\textbf{Baselines}. Briefly explain SA packages, including high T, low T, steps, etc. 

\textbf{Training and Testing}. Our approach consists of running the model for \emph{N} epochs, and then use the Sequence Pair generated by the RL agent as a starting point for SA that is further run for a certain number of steps. In our preliminary experiments, we show that SA is able to achieve a more compact placement for the same number of steps,  when initialized by our algorithm as compared to a random initialization.
For the ami49 problem, the framework is trained for 10 epochs. The agent is trained for 200 steps followed by 5000 SA steps. For the lattice problem experiment, we run our framework for 15 epochs. In each epoch, the RL agent takes 200 steps and the SA takes 5000 steps. After training, the Sequence Pair generated by the RL is used as initialization for SA. For comparison, we run SA for the same number of steps for both types of initialization. Every epoch of training takes an average of 17 seconds. Once the model has been trained, the RL initialization combined with the simulated annealing part takes an average of another 17 seconds to find the final solution for 5000 steps. In contrast, pure Simulated Annealing takes an average of 16 seconds to run for 5000 steps.

\begin{figure}[h!]
\centering
\includegraphics[width=1\textwidth]{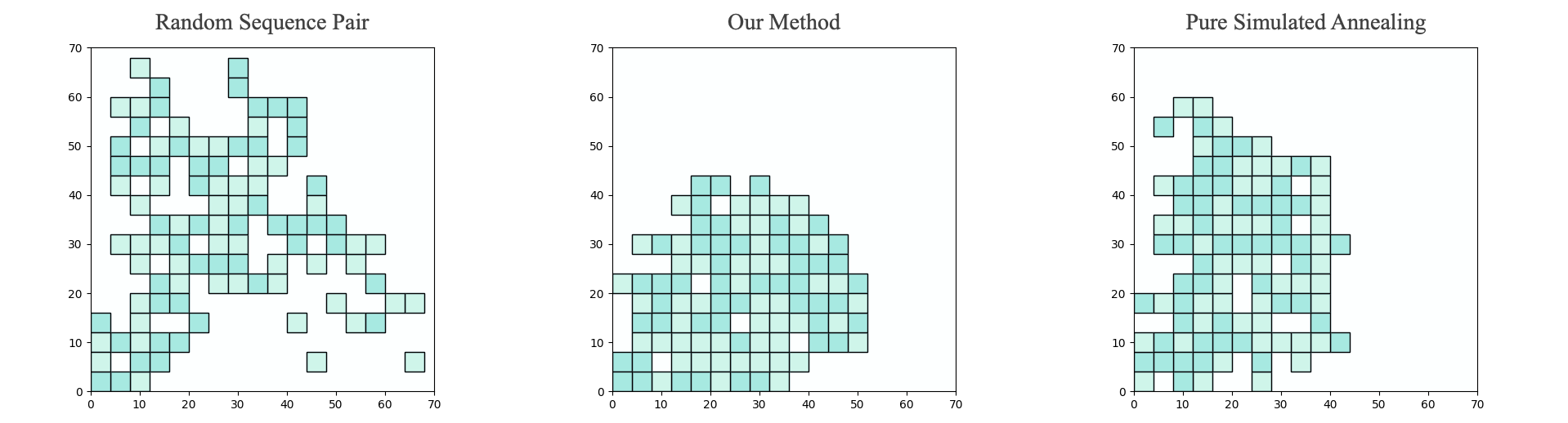}
\caption{Placements for the lattice problem after 5000 steps of Simulated Annealing (SA). For the same number of SA steps, RL initialized placements (middle) are more compact than those initialized randomly (right).}
\label{lattice_Place}
\end{figure}

\begin{table}[h!]
\label{sa_result_table_1}
  \caption{Average cost (in \textmu m) of 10 experiments on Lattice block list }
  \label{sa-table_1}
  \centering
  \begin{tabular}{lll}
    \toprule
    Number of Blocks     &  RL Initialization     & Random Initialization \\
    \midrule
    % 100 & 3,180  & 3,536     \\
    100 & 3,386  & 3,532     \\
    225     & 14,348 & 14,844   \\
    400     & 38,060 & 38,612  \\
    625    &  85,329 & 86,064\\
    \bottomrule
  \end{tabular}
\end{table}
\subsection{Results}
\label{others}
% This section describes our findings on applying the cyclic RL framework on the Lattice and ami49 <cite> \textbf{(TODO D/H)} benchmarks. Both netlists/blocklists are different in terms of the dimensions of the blocks and the cost function. The former consists of blocks of equal width and height and its cost function is focused on reducing the wirelength while the latter has blocks of different dimensions and its cost functions is focused on reducing the area of the final placement. 
% Our approach consists of running the model for \emp{N} epochs, and then use the Sequence Pair generated by the RL agent as a starting point for SA that is further run till convergence<to be changed>

% \subsection{Relative Efficiency}
Preliminary results reveal that the agent is able to outperform the baseline on both benchmarks (lattice, ami49). Table~\ref{sa-table_1} summarizes the average costs of solutions from both methods on the lattice problems of different size. For all the lattice problems with different size, our proposed methods achieved lower average cost compared with the baseline pure SA with random initialization. For $100$ blocks lattice problem, the standard deviation for RL Initialization is 146 and that of Random Initialization is 133. 

\begin{figure}[h!]
\centering
\includegraphics[width=1\textwidth]{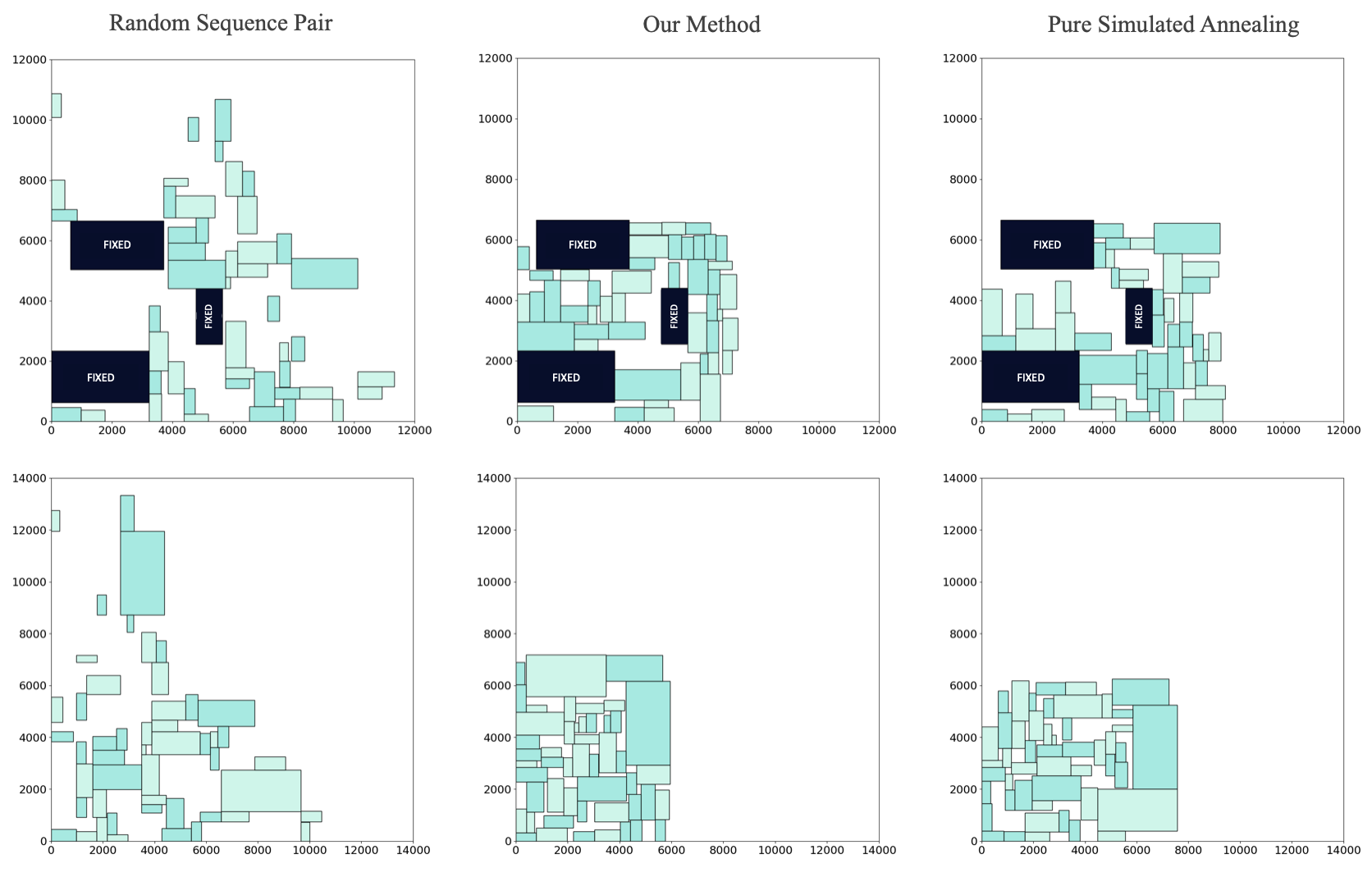}
\caption{Placements for ami49 benchmark. 3 blocks are assumed fixed in the top row. All blocks are free in the bottom row. For the same number of SA steps, RL initialized placements (middle) are more compact than those initialized randomly (right).}
\label{ami_place}
\end{figure}

The ami49 results are shown in Fig.~\ref{ami_place} and Table~\ref{ami_table_label}. Among the results, Fig.~\ref{ami_place} presents the final placement and Table~\ref{ami_table_label} compares the average cost of solutions from both the methods. The ami49 benchmark has two configurations, one involves three fixed blocks and the other has no fixed blocks as shown in Fig.~\ref{ami_place}. The fixed blocks one (Fig.~\ref{ami_place} top row) has pre-specified coordinates that need to be maintained in the final placement. This makes the learning problem more challenging as the agent tries to make perturbations that do not change the specified positions of the fixed blocks. However, the results shown in Fig.~\ref{ami_place} and Table~\ref{ami_table_label} reveal that the agent is still able to outperform the baseline (pure SA with random initialization). In this configuration, the standard deviation for RL Initialization is 1.46 and that of Random Initialization is 1.97, indicating small variance of results and actual improvement of RL Initialization. The second configuration (Fig.~\ref{ami_place} bottom row) consisting of no fixed blocks and the results also show our method's improvement from the baseline. In this case, the standard deviation for RL Initialization is 1.41 and that of Random Initialization is 1.93, indicating small variance of results and actual improvement of RL Initialization.\hfill\break

\begin{table}[h!]
\label{ami_table}
  \caption{Average cost (in mm\textsuperscript{2}) of 10 experiments on ami49 block list }
  \label{ami_table_label}
  \centering
  \begin{tabular}{lll}
    \toprule
    Type of Blocks     &  RL Initialization     & Random Initialization  \\
    \midrule
    w/ Fixed Blocks & 49  & 53    \\
    w/o Fixed Blocks     & 43 & 46    \\
   
    \bottomrule
  \end{tabular}
\end{table}
\section{Conclusion}
We presented a method to solve the placement problem by combining RL and SA in a cyclic fashion. The problem is represented in the form of Sequence Pair realized in our customized environment. By cyclic application of RL and SA, our method is able to provide a good initialization for SA and thus results in lower cost placement design by a fair margin. This shows that such a cyclic framework has the potential to achieve state of the art results. We plan to achieve such results by further formulating novel RL architectures and training strategy. Our key focus is to improve the generalization ability of our method among any random layouts of the same design as well as among totally different designs like xerox[25] and IBM[26]. Different benchmarks focus on optimizing over additional parameters such as routing congestion or DRCs. We are planning to open source the code as well. 

\newpage
\section*{References}

% References follow the acknowledgments. Use unnumbered first-level heading for
% the references. Any choice of citation style is acceptable as long as you are
% consistent. It is permissible to reduce the font size to \verb+small+ (9 point)
% when listing the references.
% {\bf Note that the Reference section does not count towards the eight pages of content that are allowed.}
% \medskip

\small
[1] Mirhoseini, Azalia, et al. "Chip Placement with Deep Reinforcement Learning." arXiv preprint arXiv:2004.10746 (2020).

[2] Markov, Igor L., Jin Hu, and Myung-Chul Kim. "Progress and challenges in VLSI placement research." Proceedings of the IEEE 103.11 (2015): 1985-2003.

[3] Brenner, Ulrich, Markus Struzyna, and Jens Vygen. "BonnPlace: Placement of leading-edge chips by advanced combinatorial algorithms." IEEE Transactions on Computer-Aided Design of Integrated Circuits and Systems 27.9 (2008): 1607-1620.

[4] Luo, Tao, and David Z. Pan. "DPlace2. 0: A stable and efficient analytical placement based on diffusion." 2008 Asia and South Pacific Design Automation Conference. IEEE, 2008.

[5] Srinivasan, Arvind, Kamal Chaudhary, and Ernest S. Kuh. "RITUAL: A performance driven placement algorithm for small cell ICs." 1991 IEEE International Conference on Computer-Aided Design Digest of Technical Papers. IEEE, 1991.

[6] Spindler, Peter, Ulf Schlichtmann, and Frank M. Johannes. "Kraftwerk2—A fast force-directed quadratic placement approach using an accurate net model." IEEE Transactions on Computer-Aided Design of Integrated Circuits and Systems 27.8 (2008): 1398-1411.

[7] Sheng, Yiqiang, Atsushi Takahashi, and Shuichi Ueno. "Relay-race algorithm: A novel heuristic approach to VLSI/PCB placement." 2011 IEEE Computer Society Annual Symposium on VLSI. IEEE, 2011.

[8] Shunmugathammal, M., C. Christopher Columbus, and S. Anand. "A novel B* tree crossover-based simulated annealing algorithm for combinatorial optimization in VLSI fixed-outline floorplans." Circuits, Systems, and Signal Processing 39.2 (2020): 900-918.

[9] Ho, Shinn-Ying, et al. "An orthogonal simulated annealing algorithm for large floorplanning problems." IEEE Transactions on Very Large Scale Integration (VLSI) Systems 12.8 (2004): 874-877.

[10] Cai, Qingpeng, et al. "Reinforcement Learning Driven Heuristic Optimization." arXiv preprint arXiv:1906.06639 (2019).

[11] Li, Zhuwen, Qifeng Chen, and Vladlen Koltun. "Combinatorial optimization with graph convolutional networks and guided tree search." Advances in Neural Information Processing Systems. 2018.

[12] Drori, Iddo, et al. "Learning to Solve Combinatorial Optimization Problems on Real-World Graphs in Linear Time." arXiv preprint arXiv:2006.03750 (2020).

[13] Schulman, John, et al. "Proximal policy optimization algorithms." arXiv preprint arXiv:1707.06347 (2017).

[14] Breuer, Melvin A. "A class of min-cut placement algorithms." Proceedings of the 14th Design Automation Conference. 1977.

[15] Dunlop, Alfred E., and Brian W. Kernighan. "A procedure for placement of standard cell VLSI circuits." IEEE Transactions on Computer-Aided Design 4.1 (1985): 92-98.

[16] Agnihotri, Ameya R., Satoshi Ono, and Patrick H. Madden. "Recursive bisection placement: Feng Shui 5.0 implementation details." Proceedings of the 2005 international symposium on Physical design. 2005.

[17] Lu, Jingwei, et al. "ePlace: Electrostatics-based placement using fast fourier transform and Nesterov's method." ACM Transactions on Design Automation of Electronic Systems (TODAES) 20.2 (2015): 1-34.

[18] Cheng, Chung-Kuan, et al. "Replace: Advancing solution quality and routability validation in global placement." IEEE Transactions on Computer-Aided Design of Integrated Circuits and Systems 38.9 (2018): 1717-1730.

[19] H.Murata,K.Fujiyoshi,M Kaneko. "VLSI/PCB placement with obstacles based on sequence pair." IEEE Transactions on Computer-Aided Design of Integrated Circuits and Systems Volume 17 Issue1 1998

[20] John Schulman, Philipp Moritz, Sergey Levine, Michael I. Jordan, and Pieter Abbeel. Highdimensional continuous control using generalized advantage estimation. International Conference on Learning Representations, 2016

[21] John Schulman, Filip Wolski, Prafulla Dhariwal, Alec Radford, and Oleg Klimov. Proximal Policy Optimization algorithms. arXiv preprint arXiv:1707.06347, 2017.

[22] Shunmugathammal, M., C. Christopher Columbus, and S. Anand. "A novel B* tree crossover-based simulated annealing algorithm for combinatorial optimization in VLSI fixed-outline floorplans." Circuits, Systems, and Signal Processing 39.2 (2020): 900-918.\hfill\break

[23] https://s2.smu.edu/~manikas/Benchmarks/Block/ami49.yal

[24] https://s2.smu.edu/~manikas/Benchmarks/Block/xerox.yal

[25] https://github.com/perrygeo/simanneal

[26] http://vlsicad.eecs.umich.edu/BK/ISPD02bench
% [1] Alexander, J.A.\ \& Mozer, M.C.\ (1995) Template-based algorithms for
% connectionist rule extraction. In G.\ Tesauro, D.S.\ Touretzky and T.K.\ Leen
% (eds.), {\it Advances in Neural Information Processing Systems 7},
% pp.\ 609--616. Cambridge, MA: MIT Press.

% [2] Bower, J.M.\ \& Beeman, D.\ (1995) {\it The Book of GENESIS: Exploring
%   Realistic Neural Models with the GEneral NEural SImulation System.}  New York:
% TELOS/Springer--Verlag.

% [3] Hasselmo, M.E., Schnell, E.\ \& Barkai, E.\ (1995) Dynamics of learning and
% recall at excitatory recurrent synapses and cholinergic modulation in rat
% hippocampal region CA3. {\it Journal of Neuroscience} {\bf 15}(7):5249-5262.

\end{document}